\documentclass[runningheads]{llncs}
\usepackage{amsmath}
\usepackage{graphicx}
\usepackage{multirow} % Required for multirow cells
\usepackage{float}
\usepackage{booktabs}
\usepackage{tabularx}
\newcolumntype{Y}{>{\centering\arraybackslash}X} 
\usepackage[colorlinks=true, linkcolor=black]{hyperref}
\usepackage{footmisc}
\begin{document}
\title{DocTabQA: Answering Questions from Long Documents Using Tables}
\newcommand*\samethanks[1][\value{footnote}]{\footnotemark[#1]}
\author{Haochen Wang\inst{1} \and
Kai Hu\inst{2} \and
Haoyu Dong\inst{3} \and
Liangcai Gao\inst{1}
}
\authorrunning{Wang et al.}
\institute{Peking University, Beijing, China \and
University of Science and Technology of China, Hefei, China \and
Microsoft Corporation\\
\email{wanghaochen326@stu.pku.edu.cn, hk970213@mail.ustc.edu.cn} \\
\email{hadong@microsoft.com, gaoliangcai@pku.edu.cn}
}
\maketitle              % typeset the header of the contribution

\def \OurTask {DocTabQA}
\def \OurDataset {QTabA}
\def \OurMethod {DocTabTalk}
\def \OurRAG {AlignLLaMA}
\def \OurPrompt {TabTalk}

\begin{abstract}
We study a new problem setting of question answering (QA), referred to as \OurTask{}. Within this setting, given a long document, the goal is to respond to questions by organizing the answers into structured tables derived directly from the document's content. Unlike traditional QA approaches which predominantly rely on unstructured text to formulate responses, \OurTask{} aims to leverage structured tables as answers to convey information clearly and systematically, thereby enhancing user comprehension and highlighting relationships between data points. To the best of our knowledge, this problem has not been previously explored. In this paper, we introduce the \OurDataset{} dataset, encompassing 300 financial documents, accompanied by manually annotated 1.5k question-table pairs. Initially, we leverage Large Language Models (LLMs) such as GPT-4 to establish a baseline. However, it is widely acknowledged that LLMs encounter difficulties when tasked with generating intricate, structured outputs from long input sequences. To overcome these challenges, we present a two-stage framework, called \OurMethod{}, which initially retrieves relevant sentences from extensive documents and subsequently generates hierarchical tables based on these identified sentences. \OurMethod{} incorporates two key technological innovations: \OurRAG{} and \OurPrompt{}, which are specifically tailored to assist GPT-4 in tackling \OurTask{}, enabling it to generate well-structured, hierarchical tables with improved organization and clarity. Comprehensive experimental evaluations conducted on both \OurDataset{} and RotoWire datasets demonstrate that our \OurMethod{} significantly enhances the performances of the GPT-4 in our proposed \OurTask{} task and the table generation task. The code and dataset are available at \url{https://github.com/SmileWHC/DocTabQA} for further research.

\keywords{Question Answering \and Table Generation \and Large Language Model \and Retrieval Augmented Generation \and Dataset}

\end{abstract}

\section{Introduction}
% Kai revised
The task of question answering (QA) has long been a cornerstone in the field of information retrieval \cite{kwok2001scaling} and natural language processing (NLP), serving as a fundamental way for machines to demonstrate their understanding of human language. At its core, QA systems aim to automatically answer questions posed by humans, typically by locating and presenting the relevant information extracted from a given text. Over the years, QA has evolved from simple factoid-based questions \cite{iyyer2014neural} to more complex, context-dependent questions \cite{choi2018quac} that require deep understanding and reasoning over extensive documents. Despite the significant advancements in language models and information retrieval techniques, effectively extracting and presenting answers from lengthy and dense documents remains a challenging endeavor. As we enter an era where data is increasingly vast and complex, there is a pressing need for QA systems to not only understand the content of documents but also to structure the retrieved information in a way that is both accessible and informative for end-users. In this context, we introduce a new problem setting of QA, referred to as \OurTask{}, which revolutionizes the output format of QA tasks by transforming textual responses into structured tables, thereby enhancing the clarity and usability of the extracted information for decision-making processes.

\begin{figure}[t]
  \centering
  \includegraphics[width=\textwidth]{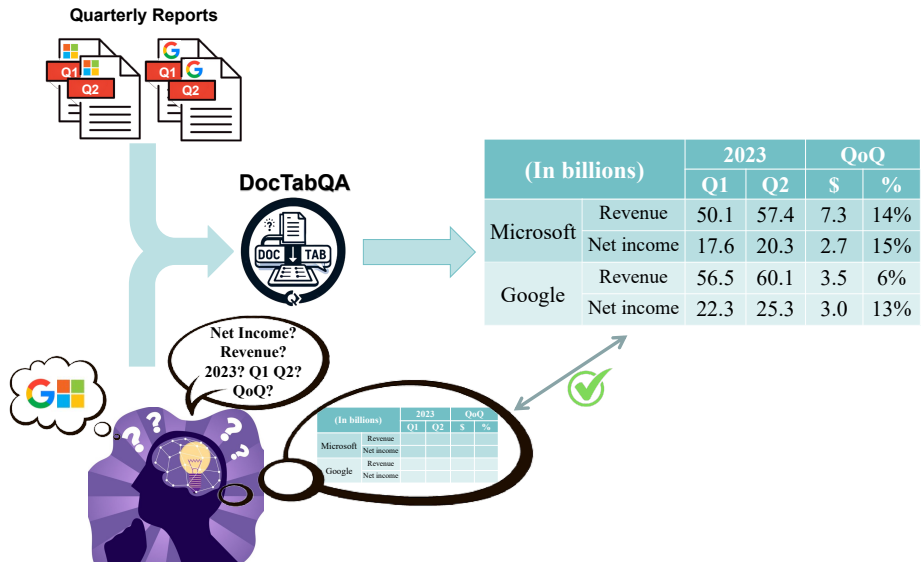}
  \caption{An illustrative example of \OurTask{}. Input documents: 2023 Q1 and Q2 quarterly reports from Microsoft and Google. Input question: ``Could you compare the revenue and net income of Microsoft and Google for Q1 and Q2 of 2023, and provide the Quarter-over-Quarter growth in both dollar amounts and percentages?".}
  \label{fig:example_table}
\end{figure}

While QA systems have greatly diversified in terms of content input, ranging from short-text snippets \cite{rajpurkar-etal-2018-know,fan-etal-2019-eli5,quiz-lelkes} to long documents \cite{NarrativeQA}, and from purely textual data to mixed media such as images (VQA \cite{antol2015vqa}, ChartQA \cite{ahmed2023realcqa}, TextVQA \cite{singh2019towards}), document images (DocVQA \cite{mathew2021docvqa}) and videos (VideoQA \cite{zhong-etal-2022-video}), the format of their outputs has remained predominantly unchanged. Traditional QA models have consistently produced answers in the form of plain text, irrespective of the complexity or the nature of the content being queried. This approach, however, often neglects the inherent structure and the relationships between pieces of information, which can be essential for users to thoroughly comprehend the context and make well-informed decisions. Structured tables, on the other hand, offer an attractive option by providing a clear and organized visual representation of data, making complex information more digestible and actionable. The advantages of table-based outputs are manifold; they can efficiently summarize key information, highlight relationships between data points, and facilitate comparisons across different dimensions. Fig.~\ref{fig:example_table} illustrates the transformation from a traditional text-based response to a structured table, showcasing how this format can encapsulate and convey intricate details more effectively.

Building upon the foundational concept of \OurTask{}, we introduce a new dataset, \OurDataset{}, to facilitate the exploration of this novel QA paradigm. This dataset is comprised of 300 financial documents, accompanied by manually annotated 1.5k question-table pairs. Unlike the datasets used in the text-to-table \cite{wu2022texttotable} task, which typically focuses on extracting key-value pairs from short input texts to construct flat tables, \OurDataset{} presents a more complex challenge. Our dataset aims to generate contextually relevant tables based on varying questions, where both the content and structure of the generated tables are dynamically controllable. Moreover, the input in our dataset is long documents that are rich in content and may describe numerous tables. The task demands not just the extraction of pertinent details but also the assembly of these details into hierarchical tables.

In this paper, we commence our exploration by utilizing Large Language Models (LLMs) such as GPT-4 \cite{openai2023gpt} to establish a baseline for our task. Our initial approach involves designing a task-specific prompt, to which we feed a (document, question, table) triple, serving as an example for in-context learning. Following this, we input the target document and question, prompting GPT-4 to generate the corresponding table as an answer. However, we observe that LLMs, including GPT-4, encounter significant challenges when tasked with generating intricate, structured outputs from lengthy input sequences. They often struggle to consistently present data in a structured format that adheres to the requirements when dealing with lengthy financial documents.

Inspired by the retrieval augmented generation (RAG) paradigm \cite{gao2023retrieval}, we present a two-stage framework, called \OurMethod{}, which initially retrieves relevant sentences from extensive documents and subsequently generates hierarchical tables based on these identified sentences. \OurMethod{} incorporates two key technological innovations: \OurRAG{} and \OurPrompt{}, which are specifically tailored to assist GPT-4 in tackling \OurTask{}, enabling it to generate well-structured, hierarchical tables with improved organization and clarity. Drawing on the query rewriting techniques \cite{ma2023query} employed in RAG, our \OurRAG{} fine-tunes a LLaMA model to rewrite input questions and document sentences to achieve semantic alignment between them. Subsequently, these rewritten questions and document sentences are leveraged to retrieve relevant text sequences. Following this, our \OurPrompt{} provides a chain-of-table generation prompt guiding the LLM through the creation of row headers, column headers, and table body cells in a sequential manner. Contrary to previous prompts that required the LLM to generate tables in one go, our \OurPrompt{} first creates the hierarchical structure of the table, including hierarchical row headers and column headers, and then fills in the content of each table body cell sequentially. Extensive experiments conducted on both \OurDataset{} and RotoWire datasets demonstrate that our \OurMethod{} significantly enhances the performance of the GPT-4 in our proposed \OurTask{} task and the table generation task.

% To overcome these challenges, we present the \OurMethod{} framework, a novel approach that incorporates two key technological innovations: \OurRAG{} and \OurPrompt{} prompts. Inspired by retrieval augmented generation (RAG) paradigm \cite{gao2023retrieval}, \OurRAG{} begins by fine-tuning a LLaMA to split each question into a series of sub-questions. These sub-questions are instrumental in pinpointing and extracting relevant sections of the source document. Subsequently, \OurRAG{} synthesizes these findings, rewriting and summarizing the sub-questions along with the retrieved information, before presenting them to GPT-4 for table generation. This allows the model to concentrate on discrete text segments, significantly bolstering the precision and pertinence of the responses. Following this, our \OurPrompt{} provides a chain-of-table generation prompt guiding the LLM through the creation of table row headers, column headers and cells in a sequential manner. Contrary to previous prompts that required the LLM to generate tables in one fell swoop, our \OurPrompt{} ensures a methodical build-up of the table's framework, thereby enhancing the clarity and precision of the output. Extensive experimental results demonstrate that our \OurMethod{} significantly enhances the performances of the GPT-4 in \OurTask{}.

The main contributions of this paper can be summarized as follows:
\begin{itemize}
  \item We are the first to present the \OurTask{} task, designed to answer questions derived from long documents using tables.
  \item To support research in this emerging question-answering paradigm, we introduce a new dataset, called \OurDataset{}, which consists of 300 financial documents, accompanied by manually annotated 1.5k question-table pairs.
  \item To address this challenge, we introduce a novel two-stage framework, named \OurMethod{}. \OurMethod{} marries two critical technological advancements: \OurRAG{} and \OurPrompt{}, which improve the capabilities of the GPT-4 in executing our proposed \OurTask{} task and the table generation task.
  % \item Our comprehensive experimental results demonstrate that \OurMethod{} is adept at generating well-organized, hierarchical structured tables from long documents, providing precise answers to questions posed by users.
\end{itemize}

\section{Related Work}

\subsection{Question Answering}
In the realm of Natural Language Processing (NLP), question answering (QA) stands as one of the most pivotal research areas and has been extensively studied for many years. This domain focuses on developing systems capable of providing precise answers to questions posed by users, drawing from a variety of underlying content sources. Traditionally, QA tasks have been categorized based on the nature of their content inputs, encompassing several distinct forms such as short-document QA \cite{rajpurkar-etal-2018-know,fan-etal-2019-eli5,quiz-lelkes}, long-document QA \cite{NarrativeQA}, Knowledge Graph QA (KGQA) \cite{huang2019knowledge}, TableQA \cite{zhu-etal-2021-tat}, Visual QA (VQA) \cite{antol2015vqa}, Document Visual QA (DocVQA) \cite{mathew2021docvqa}, and Video QA \cite{zhong-etal-2022-video}. These diverse QA tasks, each addressing a unique set of challenges and complexities, have given rise to a rich body of literature \cite{pandya2021question,yani2021challenges,jin2022survey,zou2020survey,sun2021video}. However, despite the extensive exploration of these areas, previous methodologies have consistently yielded answers in plain text format, regardless of the complexity or nature of the content in question. This conventional approach often falls short in structuring the retrieved information in an accessible and informative manner for end-users. To bridge this gap, we propose \OurTask{}, a novel paradigm that transcends the limitations of text-based responses by utilizing structured tables to answer questions derived from long documents.

\subsection{Table Generation}

Generating structured tables from textual data has garnered considerable attention in the field of information extraction \cite{wu2022texttotable,li2023sequencetosequenceset,pietruszka2022stable}. This research direction involves extracting information from unstructured text and presenting it in tabular form, thereby facilitating more accessible data interpretation and utilization. Wu et al. \cite{wu2022texttotable} approached the text-to-table conversion as a seq2seq problem, conceptualizing it as the reverse of the table-to-text generation task \cite{liu2018table}. Subsequent research has built upon this foundation, with studies \cite{he2023revisiting,rossiello2023knowgl,whitehouse2023webie,pietruszka2022stable} employing pre-trained language models, including BART \cite{lewis2019bart} and T5 \cite{raffel2023exploring}, for the generation of tables from the text. Recent studies \cite{ni2023unified,tang2023strucbench} have explored the effectiveness of LLMs in generating structured outputs. Ni et al. \cite{ni2023unified} demonstrated the use of LLMs for information extraction, specifically in the generation of key-value pairs within a structured context. Tang et al. \cite{tang2023strucbench} provided a comparative analysis of various LLMs regarding their ability to generate complex structured data. Despite the significant progress made by previous text-to-table methodologies, these approaches are predominantly data-driven, extracting information to generate tables in an uncontrollable manner, which can impede their applicability in downstream tasks. Furthermore, prior methods have typically focused on inputs consisting of just a few sentences rather than entire documents, and the output tables are often simple, lacking the complexity of hierarchical structures that can represent multi-layered relationships within the data. To address these limitations, we introduce the \OurTask{} task, which requires the generation of complex tables to answer user queries based on the input of long documents. The novelty of \OurTask{} lies in its ability to produce tables that are not only more controllable but also inherently more usable for end users.

\subsection{Retrieval Augmented Generation}
RAG (Retrieval-Augmented Generation), introduced by Lewis et al. \cite{lewis2020retrieval}, has emerged as a novel paradigm within the domain of Large Language Models (LLMs), significantly enhancing generative tasks. RAG specifically includes an initial retrieval step where LLMs query an external dataset to obtain pertinent information before commencing question-answering or text generation. The retrieval phase involves various research directions, such as enhancing semantic representations of chunks \cite{li2023angle,dai2022promptagator}; aligning queries and documents \cite{ma2023query,wang2023query2doc} and aligning retriever and LLMs \cite{yu2023augmentation}. Inspired by the query rewriting techniques \cite{ma2023query}, which aim to align the semantics between a query and its corresponding document, we introduce \OurRAG{}, which fine-tunes a LLaMA model to rewrite input questions and document sentences to achieve semantic alignment, thereby enhancing the precision of the retrieval process.

\subsection{Prompt Engineering}
A prompt is a textual instruction that delineates the task an AI is expected to execute \cite{radford2019language}. The landscape of prompting strategies is diverse, encompassing methods such as chain-of-thought \cite{wei2023chainofthought}, least-to-most \cite{zhou2022least}, and decomposed prompting \cite{khot2022decomposed}. The chain-of-thought technique enables LLMs to navigate through a sequence of intermediate steps, thereby constructing a pathway to the final answer. Diverging from this, the least-to-most approach incrementally addresses problems, starting from the simplest to the most complex, culminating in the resolution of the entire question. Decomposed prompting, distinct from the aforementioned strategies, does not confine the decomposition of tasks to a linear difficulty gradient; instead, it iteratively generates subsequent steps that various systems can implement. While much of the existing research concentrates on reasoning to address QA challenges, our focus deviates to the domain of table generation. Our \OurPrompt{} prompting strategy utilizes a chain-of-table generation prompt that first constructs the structure of the table, and then extracts relevant information to populate into the table body cells, significantly enhancing the accuracy of the generated tabular representations.

\section{\OurTask{}}

\subsection{Problem Definition}
As shown in Fig.~\ref{fig:example_table}, given a long document as the input content for QA, the goal of \OurTask{} is to answer users' specific questions by employing structured tables, thereby enhancing the clarity and usability of the extracted information. 
% Depending on the variety of queries, a single document may lead to the creation of numerous tables. Conversely, it is also possible to query information across multiple documents and compile it into a comprehensive table.

\begin{figure}[t]
  \centering
  \includegraphics[width=\textwidth]{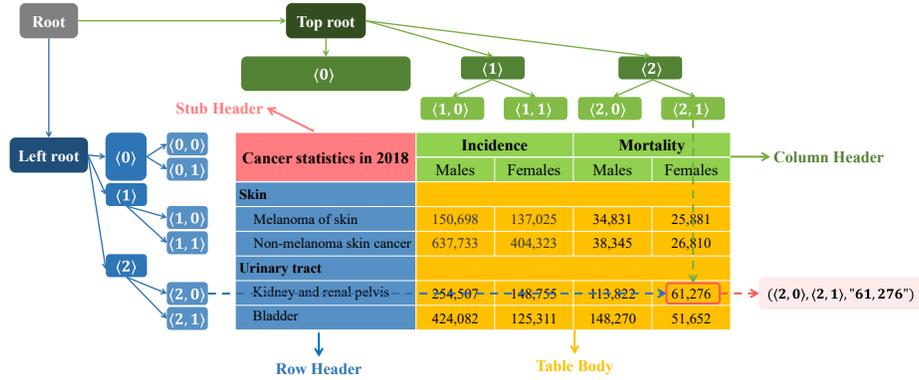}
  \caption{An example of a hierarchical table using bi-dimensional tree coordinates.}
  \label{fig:hierarchical_table_example}
\end{figure}

Formally, the input comprises a question, represented as $Q$, and a long document, denoted as $D=\{s_1, s_2, \ldots, s_{|D|}\}$, where each $s_{i}$ represents a sentence within the document $D$. The desired output is a single table, represented as $T$. Unlike previous text-to-table tasks \cite{wu2022texttotable}, which focused solely on generating \emph{flat} tables, our \OurTask{} necessitates the creation of \emph{hierarchical} tables that are tailored to represent the intricate hierarchical relationships among the data. Following \cite{wang2021tuta}, we use the \emph{bi-dimensional coordinate tree} to structure and systematically define cell location in generally structured tables considering both spatial and hierarchical information. As shown in Fig.~\ref{fig:hierarchical_table_example}, each hierarchical table consists of a stub header, a hierarchical row header, a hierarchical column header, and a table body. The hierarchical row header forms the \emph{left coordinate tree} of the bi-dimensional coordinate system, while the hierarchical column header constitutes the \emph{top coordinate tree}. Each cell within the table body can precisely pinpoint its location using a unique set of bi-dimensional tree coordinates. For example, the bi-dimensional tree coordinates for the cell ``61,276" is $(\langle 2, 0 \rangle, \langle 2, 1 \rangle)$, where $\langle 2, 0 \rangle$ is the left tree coordinate and $\langle 2, 1 \rangle$ is the top tree coordinate. Therefore, each cell within Table $T$ is defined by a triplet: $(\langle left\_tree\_coord \rangle, \langle top\_tree\_coord \rangle, text)$. The goal of \OurTask{} is to generate such a hierarchical table $T$ as an answer, given an input document $D$ and a question $Q$.

\subsection{\OurDataset{} Dataset}

We introduce a new dataset, \OurDataset{}, to facilitate the research of our proposed \OurTask{}. This section delineates the procedures for document collection and the annotation pipeline employed in the construction of the dataset.

\subsubsection{Document Collection.} 
To construct the \OurDataset{} dataset, we download approximately 300 financial reports issued over the past two years from the Securities and Exchange Commission (SEC)\footnote{https://www.sec.gov/} in PDF format. Subsequently, we utilize the Adobe PDF Services API\footnote{https://developer.adobe.com/document-services/apis/pdf-services/} to extract both text and table contents from the reports for further processing within our annotation pipeline.

\begin{figure}[t]
  \centering
  \includegraphics[width=0.8\textwidth]{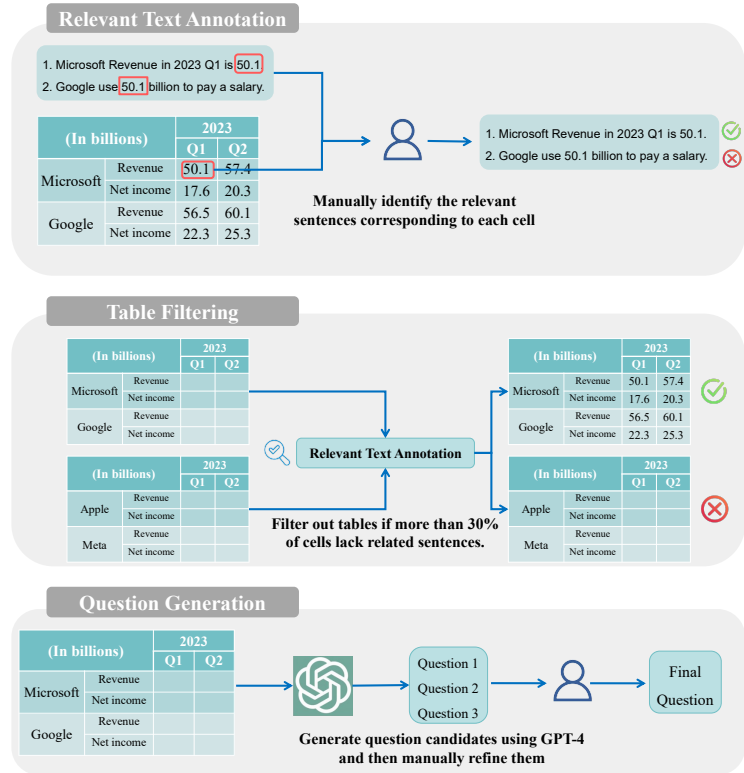}
  \caption{\OurDataset{} dataset annotation pipeline.}
  \label{fig:annotation_pipeline}
\end{figure}

\subsubsection{Annotation Pipeline.}
As shown in Fig.~\ref{fig:annotation_pipeline}, given the downloaded text contents and the corresponding tables, we manually annotate the (document, question, table) triples utilizing the following pipeline:
\begin{itemize}
  \item \textbf{Relevant Text Annotation:} The preliminary step in our annotation pipeline for each table involves the identification of document sentences that are pertinent to the table's description. This task is initiated by deploying regular expressions to accurately locate sentences that contain text matching the content in the cells of the table. However, it is possible for multiple sentences to match the content within a specific cell. In these cases, a meticulous manual review of these sentences is conducted to ascertain their relevance to the table in question. 
  %Our screening process encompassed approximately 8k cells and roughly 70k sentences, culminating in the creation of a dataset of exceptional quality. The corpus of annotated sentences is instrumental for assessing the retrieval stage of our question-answering system.
  \item \textbf{Table Filtering:} Nevertheless, certain cells within a table may not have corresponding descriptive sentences in the document. In scenarios where 30\% or more of a table's cells are devoid of associated descriptive statements, we exclude the table from our dataset. This measure is taken to ensure the high quality and integrity of the data.
  \item \textbf{Question Generation}: Finally, we employ GPT-4 to generate an initial set of questions based on the input table. Detailed prompting instructions are provided in Appendix A.1. Subsequently, these questions undergo a manual refinement process to ensure their direct alignment with the table's content.
\end{itemize}

\subsubsection{Dataset Statistics.} Detailed statistical comparisons between our \OurDataset{} dataset and existing text-to-table datasets are provided in Table.~\ref{tab:dataset_statistics}. As indicated in Table.~\ref{tab:dataset_statistics}, the input text sequences in our \OurDataset{} consist of long documents, which are substantially lengthier than those found in previous datasets. In addition, the tables generated by our \OurDataset{} are significantly more complex than those in prior text-to-table datasets. Previous datasets typically generated tables with only two columns, primarily consisting of key-value pair tables where each row corresponds to an individual key-value pair. Although the RotoWire dataset is much larger, it contains only flat tables. In contrast, tables from \OurDataset{} feature numerous hierarchical structures.

\begin{table}[t]
\centering
\caption{Detailed statistical comparisons among \OurDataset{} dataset and existing text-to-table datasets.}
\label{tab:dataset_statistics}
\begin{tabularx}{\textwidth}{lYYYYcccc}
\toprule
\multirow{2}{*}{Dataset} & \multirow{2}{*}{Train} & \multirow{2}{*}{Val} & \multirow{2}{*}{Test} & \multirow{2}{*}{\shortstack{Input \\ Tokens}} && \multicolumn{3}{c}{Output Tables} \\ \cline{7-9}
 &  &  &  & && Rows \& Columns & Flat & Hierarchical \\
 \midrule
E2E \cite{nayak2019effective} & 42.1k & 4.7k & 4.7k & 24.9 && 4.6 \& 2.0 & 49.8k & 0 \\
WikiTableText \cite{bao2018tabletotext} & 10.0k & 1.3k & 2.0k & 20.0 && 4.3 \& 2.0 & 12.0k & 0 \\
WikiBio \cite{lebret2016neural} & 582.7k & 72.8k & 72.7k & 122.3 && 4.2 \& 2.0 & 655.4k & 0 \\
RotoWire \cite{wiseman-etal-2017-challenges} & 3.4k & 727 & 728 & 373.7 && 7.3 \& \textbf{8.8} & 4.8k & 0 \\
\OurDataset{} & 1.4k & - & 160 & \textbf{19.4k} && \textbf{8.4} \& 4.1 & 1k & \textbf{532} \\
\bottomrule
\end{tabularx}
\end{table}

\section{Methodology}

% kai revised
In this paper, we introduce a novel two-stage framework, dubbed \OurMethod{}, designed to tackle the \OurTask{} task. As depicted in Fig.~\ref{fig:method}, the architecture of \OurMethod{} comprises two core components: (1) An \OurRAG{} module aligns the input questions and document sentences to efficiently retrieve relevant information from extensive texts based on the queries; (2) A new prompting strategy, termed \OurPrompt{}, crafted to generate accurate and hierarchically structured tables from the extracted relevant sentences. In the following sections, we provide an in-depth exploration of both the \OurRAG{} and \OurPrompt{} components.

\begin{figure}[t]
\includegraphics[width=\textwidth]{Figures/overview_v1.pdf}
\caption{Overview of \OurMethod{}. } \label{fig:method}
\end{figure}

\subsection{\OurRAG{}}

Inspired by the query rewriting techniques \cite{ma2023query} employed in RAG, we fine-tune a LLaMA model, which we named \OurRAG{}, to rewrite input questions and document sentences for semantic alignment. Specifically, \OurRAG{} initially decomposes each input question into a set of sub-questions. These sub-questions are then leveraged to perform multiple parallel retrievers to retrieve relevant document sentences. This strategy proves advantageous when addressing complex questions comprising multiple sub-problems. Additionally, we utilize \OurRAG{} to rewrite document sentences so that all sentences are structured with specific data as the subject, which makes them more akin to answers to the question, thereby enhancing the accuracy and recall of the retrieval process. Finally, we utilize a Sentence-BERT model \cite{reimers2019sentence} as our retrieval mechanism, calculating the cosine similarity between each rewritten sub-question and the rewritten document sentences. We then extract the top-$K$ sentences with the highest similarity measures, identifying them as the most relevant sentences and feeding them into our table generation stage.

\subsection{\OurPrompt{}}

\begin{figure}[t]
  \centering
  \includegraphics[width=1\textwidth]{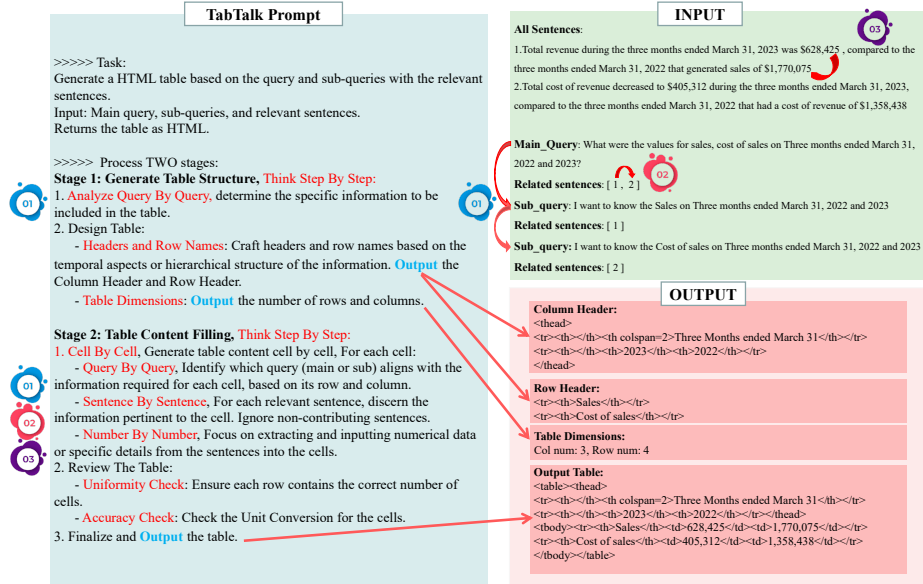}
  \caption{A schematic view of the proposed \OurPrompt{} prompt.} \label{fig:prompt}
\end{figure}

During the table generation phase, we introduce a novel prompting mechanism dubbed \OurPrompt{}. This approach is designed to accurately synthesize hierarchically structured tables by interpreting input questions and extracted relevant sentences. As illustrated in Fig.~\ref{fig:prompt}, \OurPrompt{} systematically deconstructs the intricate process of table generation into a two-stage operation. In the initial stage, \OurPrompt{} focuses on the hierarchical architecture of the table, establishing both the hierarchical row and column headers, thereby setting up a \emph{bi-dimensional tree coordinates} for the table. In the subsequent stage, \OurPrompt{} fills the table body with data, aligning the cell contents with the information presented in the sentences to finalize the construction of the table. This strategy not only facilitates the model's thought process and the accumulation of information in an orderly progression from simple to complex but also encourages self-reflection at each step of output. This ensures that the resulting table structure is complete and the content is accurate. 

\subsubsection{Table Structure Generation.}

During the table structure generation stage, we employ a chain-of-thought \cite{wei2023chainofthought} approach to progressively delve into the process. This strategy, which progresses from sub-queries to the main query, determines the information that the table should include in a part-to-whole sequence, thereby accurately establishing the content for row headers and column headers. Subsequently, we instruct the model to sequentially produce the row headers, column headers, and dimensions for the constructed table structure. This process not only facilitates self-reflection within the model but also markedly enhances the structural precision of the table.

\subsubsection{Table Content Filling.}

During the table content filling stage, we continue to employ the chain-of-thought method, guiding the model to meticulously fill each cell in the table body. This process includes formulating precise queries, searching for relevant sentences, and verifying specific information within each sentence, such as numerical data. These steps significantly enhance the accuracy of the generated content. Before finalizing and outputting the complete table, we conduct a thorough verification to ensure the table's format is correct and that any content involving unit conversions within the table body cells is accurately executed.

\section{Experiments}

\subsection{Datasets}

We perform evaluations using a publicly accessible text-to-table dataset, RotoWire \cite{wiseman-etal-2017-challenges}, and our newly introduced dataset, \OurDataset{}, which is specifically curated for the \OurTask{} task, to ascertain the efficacy of our \OurMethod{} framework.

\textbf{RotoWire} is a dataset designed for the text-to-table conversion, comprising textual descriptions of basketball matches along with comprehensive statistical tables. For each description, two tables can be generated: one representing the team scores and the other representing individual player scores. We focus on generating the player table due to its higher complexity. This dataset has been randomly divided into training, validation, and test subsets, containing 3398, 727, and 728 text-table pairs, respectively.

\textbf{\OurDataset{}} is our proposed dataset, tailored for the \OurTask{} problem. It encompasses 300 financial reports from the past two years and is randomly partitioned into training and test subsets, with 240 and 60 documents each. For the training and test subsets, we manually annotate 1.4K and 160 question-table pairs, respectively.

\subsection{Evaluation Metrics}
Our \OurMethod{} framework operates in a two-stage manner, utilizing specific metrics for each stage to assess the performance of our proposed technologies.

During the retrieval stage, we utilize the top-$K$ recall metric to evaluate the capability of various models to retrieve and rank relevant sentences effectively.

In the table generation stage, in line with \cite{tang2023strucbench}, we decompose the similarity assessment between the two tables into structural and content components. Given that a table's structure can be represented as a bi-dimensional tree, as illustrated in Fig.~\ref{fig:hierarchical_table_example}, we propose to use the tree edit distance similarity (TEDS) as the metric for evaluating structural similarity. For assessing table content similarity, our approach diverges from previous text-to-table methodologies, such as those outlined in \cite{wu2022texttotable}, which typically assess similarity at the individual cell level within a table. Instead, we utilize the \emph{bi-dimensional tree coordinates} to transform each cell in the table body into a distinct key-value pair. Take the cell containing \textit{``61, 276"} as an example. This cell is converted into a key-value pair represented as ($\langle \textit{``Urinary tract", ``Kidney and renal pelvis"} \rangle$, $\langle \textit{``Mortality", ``Females"} \rangle$, \textit{``61, 276"}). We then measure the content similarity by comparing these key-value pairs, which are derived from the generated tables, against the key-value pairs from the ground-truth tables. To evaluate table content similarity, we employ a suite of classical metrics that are well-established for assessing textual similarity, including character n-gram F-score (Chrf) \cite{popovic-2015-chrf} and BERTScore \cite{zhang2019bertscore}. Following \cite{tang2023strucbench}, we additionally engage GPT-4 to evaluate the generated tables, tasking it with scoring the similarity of both content and structure. These evaluations are referred to as the GPT-Score. Detailed information can be found in \cite{tang2023strucbench}.

\subsection{Implementation Details}
Our \OurRAG{} utilizes the conventional instruction tuning method to fine-tune LLaMA2-7B \cite{touvron2023llama} with LoRA \cite{hu2021lora}. This instruction tuning empowers LLaMA2 to rewrite input questions and document sentences, achieving better alignment between them. For the fine-tuning process of the LLaMA-7B model, we employ a dataset comprising 1.4k tables and 10k sentences, all of which feature significant numeric information extracted from the documents of the training set. The ground truth of rewritten questions and documents in this fine-tuning dataset is generated by GPT-4. Detailed prompting instructions are provided in Appendix A.2. The fine-tuning is carried out on a workstation equipped with 2 NVIDIA Tesla A800 GPUs (80 GB of memory). In the retrieval phase, we prioritize the top-30 relevant document sentences, which are then fed into the subsequent table generation stage. During the table generation phase, our approach diverges from previous text-to-table methods that produce tables in Markdown format, which lack the capability to represent hierarchical structures. Instead, we opt to generate tables in HTML format, enabling the depiction of complex, hierarchical table structures.

\subsection{Comparisons with Prior Arts}

In our \OurMethod{} framework, there are two key technological innovations: \OurRAG{} and \OurPrompt{}. We conduct comparisons of these two modules against state-of-the-art methods respectively.

\begin{table}[t]
\centering
\caption{Performance comparison of retrieval stage on \OurDataset{}. (R@$K$ denotes the recall rate for the top-$K$ retrieved results, in \%)}
\begin{tabularx}{0.8\textwidth}{cYYYYYY}
\toprule
Model  & R@10 & R@20 & R@30 & R@40 & R@50 & R@60 \\
\midrule
MiniLM \cite{wang2020minilm} & 62.47 & 73.87 & 79.51 & 81.55 & 84.19 & 85.82 \\
MPNet \cite{song2020mpnet} & 58.89 & 72.22 & 76.72 & 81.85 & 85.3 & 88.01 \\
GPT-4 \cite{openai2023gpt} & 60.28 & 72.80 & 76.57 & 77.84 & 78.61 & 78.84 \\
\OurRAG{} & \textbf{69.64} & \textbf{80.37} & \textbf{85.22} & \textbf{88.41} & \textbf{92.01} & \textbf{94.05} \\
\bottomrule
\end{tabularx}
\label{tab:recall}
\end{table}

\subsubsection{Retrieval Stage.}
During the retrieval phase, we implement several baseline methodologies, including MiniLM \cite{wang2020minilm}, MPNet \cite{song2020mpnet}, and GPT-4 \cite{openai2023gpt}, to fetch document sentences pertinent to the posed questions. MiniLM and MPNet accomplish this by computing the semantic similarity between the queries and document sentences, thus retrieving sentences of relevance. In contrast, GPT-4 adopts in-context learning, utilizing examples to infer the relevance between the input document sentences and the query, subsequently generating a similarity score for these entities. Detailed prompting instructions are provided in Appendix A.3. Our \OurRAG{} initially rewrites both questions and document sentences before employing Sentence-BERT to assess the similarity among these modified sentences. This approach markedly enhances the top-$K$ recall. As evidenced in Table.~\ref{tab:recall}, \OurRAG{} achieves an impressive recall rate of 85.22\% at top-30, significantly surpassing other methods.

\subsubsection{Table Generation Stage.}

\begin{table}[t]
\centering
\caption{Performance comparison of table generation stage on \OurDataset{}. Metrics include BERT-Score (BERT, \%), GPT-Score (GPT, on a scale of 0-10), and Tree Edit Distance Similarity (TEDS, \%). (R\&C Header represents row header and column header)}
\begin{tabularx}{\textwidth}{llcYYcYYcYY}
\toprule
\multirow{3}{*}{\#}&\multirow{3}{*}{Model} && \multicolumn{5}{c}{Content Similarity} && \multicolumn{2}{c}{Structure Similarity} \\ \cline{4-8} \cline{10-11}
&&& \multicolumn{2}{c}{Table Body} && \multicolumn{2}{c}{R\&C Header} && \multirow{2}{*}{TEDS} & \multirow{2}{*}{GPT} \\ \cline{4-5} \cline{7-8}
&&&  BERT & GPT && BERT & GPT && & \\
\midrule
1&GPT-4 && 62.59 & 4.16 && 63.20 & 4.43 && 92.15 & 4.29 \\
2&\quad + Two Stage && 67.14 & 4.26 && 64.31 & 4.66 && 92.97 & 4.36 \\
3&\quad\quad + \OurRAG{} && 70.08 & 4.32 && 64.70 & 4.92 && 94.15 & 4.47 \\
4&\quad\quad\quad + \OurPrompt{} && \textbf{74.76} & \textbf{5.90} && \textbf{66.36} & \textbf{5.44} && \textbf{96.33} & \textbf{6.59} \\
\bottomrule
\label{tab:prompt}
\end{tabularx}
\end{table}

After retrieving relevant sentences, the goal of our table generation stage is to generate a hierarchically structured table from these sentences as an answer to the input question. We validate the effectiveness of our \OurPrompt{} on two datasets, \OurDataset{} and RotoWire.

\textbf{\OurDataset{}.} In our initial experiment, we establish a straightforward baseline by using a simple prompting strategy to assess the one-shot in-context learning capabilities of GPT-4. This involves employing a single illustrative example and a prompt that clearly outlines the task. The prompt includes a (document, question, table) triplet to guide GPT-4 toward generating tabular results after processing the entire document text and the corresponding query. Detailed prompting instructions are provided in Appendix A.4.

As shown in row \#1 of Table~\ref{tab:prompt}, the baseline approach does not perform optimally, primarily due to GPT-4's challenges with processing lengthy documents and maintaining context across extended texts. We then apply a two-stage framework as indicated in row \#2 of Table~\ref{tab:prompt}, which involves an initial retrieval phase using GPT-4 to extract relevant sentences, followed by a generation phase. The results demonstrate a significant improvement over the baseline method, underscoring the effectiveness of the two-stage strategy.

Further enhancements are observed when we introduce our \OurRAG{} module as the retrieval mechanism, as depicted in row \#3 of Table~\ref{tab:prompt}. The performance is notably better, especially when our \OurPrompt{} prompting strategy is implemented. This strategy leads to marked improvements in the generated table structures, with a 2.18\% increase in the TEDS metric, indicating that our method is more effective at generating complex hierarchical tables. Overall, the comparative analysis shows that our \OurMethod{} framework, with the integration of \OurRAG{} and \OurPrompt{} prompts, significantly enhances GPT-4's ability to structure and present information in tabular form for complex QA tasks involving long documents.

\begin{table}[t]
\centering
\caption{Performance comparison of text-to-table on RotoWise. Metrics include character n-gram F-score (Chrf, \%), BERT-Score (BERT, \%), and Tree Edit Distance Similarity (TEDS, \%). (R Header and C Header represent row header and column header, respectively. ICL represents in-context learning.)}
\begin{tabularx}{\textwidth}{llcYYcYYcYYcc}
\toprule
\multirow{3}{*}{\#}&\multirow{3}{*}{Model} && \multicolumn{8}{c}{Content Similarity} && {Structure} \\ \cline{4-11} \cline{13-13}
&&& \multicolumn{2}{c}{Table Body} && \multicolumn{2}{c}{R Header} &&  \multicolumn{2}{c}{C Header} && \multirow{2}{*}{TEDS}  \\ \cline{4-5} \cline{7-8} \cline{10-11}
&&&  Chrf & BERT && Chrf & BERT && Chrf & BERT && \\
\midrule
\multirow{3}{*}{Supervised}& Sent-level RE \cite{zhong-chen-2021-frustratingly} && 83.42 & 85.35 && 93.00 & 90.98 && 89.38 & 93.07 && - \\
& Seq2Seq-c \cite{wu2022texttotable} && 84.74 & 88.97 && 94.0 & 93.71 && 91.26 & 94.41 && - \\
& Seq2Seq\&set \cite{li2023sequencetosequenceset} && \textbf{85.75} & \textbf{90.93} && 94.48 & 96.43 && \textbf{91.60} & \textbf{95.08} && - \\
\midrule
\multirow{2}{*}{ICL} & GPT-4 && 82.48 & 82.40 && 96.06 & 97.46 && 67.62 & 77.20 && 100.0 \\
& \quad + \OurPrompt{} && \textbf{85.75} & 84.92 && \textbf{97.53} & \textbf{98.68} && 71.30 & 80.97 && 100.0 \\
\bottomrule
\label{tab:RotoWire}
\end{tabularx}
\end{table}

\textbf{RotoWire.} We apply our \OurPrompt{} to the text-to-table task, and as shown in Table.~\ref{tab:RotoWire}, our method significantly improves GPT-4's ability to generate structured tables. Our method achieves comparable results with supervised methods in the table body and even better on row headers with only a few examples for in-context learning. On column headers, our approach improves GPT-4 but still falls short compared to supervised methods, which is due to the limited examples provided by the in-context learning. Additionally, since the tables in this dataset are all simple, flat tables, the similarity in table structure for the generated tables is always 100\%.

\section{Limitations}

Despite our \OurDataset{} bringing new challenges (\OurTask{}) to the Document-based Question Answering (DocQA) domain and our \OurMethod{} showing promising results, there are still several limitations in both our dataset and approach that need to be addressed. For the dataset, its exclusive focus on English sources and single-document input restricts the breadth of our study and its potential cross-linguistic application. Expanding the dataset to accommodate multilingual content and multi-document inputs would greatly enhance the versatility and depth of the DocQA framework. Regarding our method, the initial retrieval stage is relatively simplistic, primarily assessing the relevance of document content to the question without deeply engaging the model's inferential capabilities. This superficial process may overlook complex relationships that require advanced reasoning. Therefore, improving our method's capacity for inference is crucial, ensuring that it can more accurately discern and utilize relevant information for question answering in future research.

\section{Conclusion and Future Work}

In this paper, we make a significant contribution to the field of Document-based Question Answering by introducing \OurTask{}, a novel problem setting that transforms answers from textual responses into structured tables. Through the development and evaluation of the \OurDataset{} dataset, we propose a two-stage framework, called \OurMethod{}, to improve the performance of GPT-4. \OurMethod{} incorporates two key technological innovations: \OurRAG{} and \OurPrompt{}, which are specifically tailored to assist GPT-4 in tackling \OurTask{}, enabling it to generate well-structured, hierarchical tables with improved organization and clarity.  The experimental results on the \OurDataset{} and the RotoWire dataset convincingly demonstrate the effectiveness of our method. Our approach marks a substantial step forward in presenting information succinctly and systematically.

In the future, we will aim to address the limitations highlighted in our study. We plan to expand the dataset to include multilingual documents and multi-document inputs, which will challenge and potentially improve the robustness of DocQA systems. Moreover, we will focus on enhancing the inferential capabilities of our method, enabling it to grasp the subtleties and complexities of document content beyond the superficial level. We will also work on improving the model's reasoning abilities, which are essential for accurately determining the relevance of information in response to user queries. By tackling these challenges, we expect to advance the generation of structured summaries and the overall effectiveness of question answering systems for complex, long-form documents.

\section{Acknowledgement}
This work is supported by the projects of National Science and Technology Major Project (2021ZD0113301) and National Natural Science Foundation of China (No. 62376012), which is also a research achievement of Key Laboratory of Science, Technology and Standard in Press Industry (Key Laboratory of Intelligent Press Media Technology).

\bibliographystyle{splncs04}
\bibliography{papers}

\begin{thebibliography}{10}
\providecommand{\url}[1]{\texttt{#1}}
\providecommand{\urlprefix}{URL }
\providecommand{\doi}[1]{https://doi.org/#1}

\bibitem{ahmed2023realcqa}
Ahmed, S., Jawade, B., Pandey, S., Setlur, S., Govindaraju, V.: Realcqa: Scientific chart question answering as a test-bed for first-order logic. In: ICDAR. pp. 66--83. Springer (2023)

\bibitem{antol2015vqa}
Antol, S., Agrawal, A., Lu, J., Mitchell, M., Batra, D., Zitnick, C.L., Parikh, D.: Vqa: Visual question answering. In: ICCV. pp. 2425--2433 (2015)

\bibitem{bao2018tabletotext}
Bao, J., Tang, D., Duan, N., Yan, Z., Lv, Y., Zhou, M., Zhao, T.: Table-to-text: Describing table region with natural language. In: AAAI (2018)

\bibitem{choi2018quac}
Choi, E., He, H., Iyyer, M., Yatskar, M., Yih, W.t., Choi, Y., Liang, P., Zettlemoyer, L.: {Q}u{AC}: Question answering in context. In: EMNLP. pp. 2174--2184 (2018)

\bibitem{dai2022promptagator}
Dai, Z., Zhao, V.Y., Ma, J., Luan, Y., Ni, J., Lu, J., Bakalov, A., Guu, K., Hall, K.B., Chang, M.W.: Promptagator: Few-shot dense retrieval from 8 examples. In: ICLR (2023)

\bibitem{fan-etal-2019-eli5}
Fan, A., Jernite, Y., Perez, E., Grangier, D., Weston, J., Auli, M.: {ELI}5: Long form question answering. In: ACL. pp. 3558--3567 (2019)

\bibitem{gao2023retrieval}
Gao, Y., Xiong, Y., Gao, X., Jia, K., Pan, J., Bi, Y., Dai, Y., Sun, J., Wang, H.: Retrieval-augmented generation for large language models: A survey. arXiv preprint arXiv:2312.10997  (2023)

\bibitem{he2023revisiting}
He, Y., Hu, J., Tang, B.: Revisiting event argument extraction: Can {EAE} models learn better when being aware of event co-occurrences? In: ACL. pp. 12542--12556 (2023)

\bibitem{hu2021lora}
Hu, E.J., Shen, Y., Wallis, P., Allen-Zhu, Z., Li, Y., Wang, S., Wang, L., Chen, W.: Lo{RA}: Low-rank adaptation of large language models. In: ICLR (2022)

\bibitem{huang2019knowledge}
Huang, X., Zhang, J., Li, D., Li, P.: Knowledge graph embedding based question answering. In: WSDM. pp. 105--113 (2019)

\bibitem{iyyer2014neural}
Iyyer, M., Boyd-Graber, J., Claudino, L., Socher, R., Daum{\'e}~III, H.: A neural network for factoid question answering over paragraphs. In: EMNLP. pp. 633--644 (2014)

\bibitem{jin2022survey}
Jin, N., Siebert, J., Li, D., Chen, Q.: A survey on table question answering: recent advances. In: CCKS. pp. 174--186. Springer (2022)

\bibitem{khot2022decomposed}
Khot, T., Trivedi, H., Finlayson, M., Fu, Y., Richardson, K., Clark, P., Sabharwal, A.: Decomposed prompting: A modular approach for solving complex tasks. arXiv preprint arXiv:2210.02406  (2022)

\bibitem{NarrativeQA}
Kočiský, T., Schwarz, J., Blunsom, P., Dyer, C., Hermann, K.M., Melis, G., Grefenstette, E.: {The NarrativeQA Reading Comprehension Challenge}. Trans. Assoc. Comput. Linguist.  \textbf{6},  317--328 (2018)

\bibitem{kwok2001scaling}
Kwok, C.C., Etzioni, O., Weld, D.S.: Scaling question answering to the web. In: WWW. pp. 150--161 (2001)

\bibitem{lebret2016neural}
Lebret, R., Grangier, D., Auli, M.: Neural text generation from structured data with application to the biography domain. In: EMNLP. pp. 1203--1213 (2016)

\bibitem{quiz-lelkes}
Lelkes, A.D., Tran, V.Q., Yu, C.: Quiz-style question generation for news stories. In: WWW. p. 2501–2511 (2021)

\bibitem{lewis2019bart}
Lewis, M., Liu, Y., Goyal, N., Ghazvininejad, M., Mohamed, A., Levy, O., Stoyanov, V., Zettlemoyer, L.: {BART}: Denoising sequence-to-sequence pre-training for natural language generation, translation, and comprehension. In: ACL. pp. 7871--7880 (2020)

\bibitem{lewis2020retrieval}
Lewis, P., Perez, E., Piktus, A., Petroni, F., Karpukhin, V., Goyal, N., K{\"u}ttler, H., Lewis, M., Yih, W.t., Rockt{\"a}schel, T., et~al.: Retrieval-augmented generation for knowledge-intensive nlp tasks. In: NeurIPS. pp. 9459--9474 (2020)

\bibitem{li2023sequencetosequenceset}
Li, T., Wang, Z., Shao, L., Zheng, X., Wang, X., Su, J.: A sequence-to-sequence\&set model for text-to-table generation. In: ACL-Findings. pp. 5358--5370 (2023)

\bibitem{li2023angle}
Li, X., Li, J.: Angle-optimized text embeddings. arXiv preprint arXiv:2309.12871  (2023)

\bibitem{liu2018table}
Liu, T., Wang, K., Sha, L., Chang, B., Sui, Z.: Table-to-text generation by structure-aware seq2seq learning. In: AAAI (2018)

\bibitem{ma2023query}
Ma, X., Gong, Y., He, P., Zhao, H., Duan, N.: Query rewriting for retrieval-augmented large language models. In: EMNLP. pp. 5303--5315 (2023)

\bibitem{mathew2021docvqa}
Mathew, M., Karatzas, D., Jawahar, C.: Docvqa: A dataset for vqa on document images. In: WACV. pp. 2200--2209 (2021)

\bibitem{nayak2019effective}
Nayak, T., Ng, H.T.: Effective modeling of encoder-decoder architecture for joint entity and relation extraction. In: AAAI. pp. 8528--8535 (2020)

\bibitem{ni2023unified}
Ni, X., Li, P.: Unified text structuralization with instruction-tuned language models. arXiv preprint arXiv:2303.14956  (2023)

\bibitem{openai2023gpt}
OpenAI: Gpt-4 technical report. arXiv preprint arXiv:2303.08774  (2023)

\bibitem{pandya2021question}
Pandya, H.A., Bhatt, B.S.: Question answering survey: Directions, challenges, datasets, evaluation matrices. arXiv preprint arXiv:2112.03572  (2021)

\bibitem{pietruszka2022stable}
Pietruszka, M., Turski, M., Borchmann, {\L}., Dwojak, T., Pa{\l}ka, G., Szyndler, K., Jurkiewicz, D., Garncarek, {\L}.: Stable: Table generation framework for encoder-decoder models. arXiv preprint arXiv:2206.04045  (2022)

\bibitem{popovic-2015-chrf}
Popovi{\'c}, M.: chr{F}: character n-gram {F}-score for automatic {MT} evaluation. In: WMT. pp. 392--395 (2015)

\bibitem{radford2019language}
Radford, A., Wu, J., Child, R., Luan, D., Amodei, D., Sutskever, I., et~al.: Language models are unsupervised multitask learners. OpenAI blog  \textbf{1}(8), ~9 (2019)

\bibitem{raffel2023exploring}
Raffel, C., Shazeer, N., Roberts, A., Lee, K., Narang, S., Matena, M., Zhou, Y., Li, W., Liu, P.J.: Exploring the limits of transfer learning with a unified text-to-text transformer. J Mach Learn Res  \textbf{21}(1) (jan 2020)

\bibitem{rajpurkar-etal-2018-know}
Rajpurkar, P., Jia, R., Liang, P.: Know what you don{'}t know: Unanswerable questions for {SQ}u{AD}. In: ACL. pp. 784--789 (2018)

\bibitem{reimers2019sentence}
Reimers, N., Gurevych, I.: Sentence-{BERT}: Sentence embeddings using {S}iamese {BERT}-networks. In: EMNLP-IJCNLP. pp. 3982--3992 (2019)

\bibitem{rossiello2023knowgl}
Rossiello, G., Chowdhury, M.F.M., Mihindukulasooriya, N., Cornec, O., Gliozzo, A.M.: Knowgl: Knowledge generation and linking from text. In: AAAI. pp. 16476--16478 (2023)

\bibitem{singh2019towards}
Singh, A., Natarjan, V., Shah, M., Jiang, Y., Chen, X., Parikh, D., Rohrbach, M.: Towards vqa models that can read. In: CVPR. pp. 8317--8326 (2019)

\bibitem{song2020mpnet}
Song, K., Tan, X., Qin, T., Lu, J., Liu, T.Y.: Mpnet: Masked and permuted pre-training for language understanding. In: NeurIPS. pp. 16857--16867 (2020)

\bibitem{sun2021video}
Sun, G., Liang, L., Li, T., Yu, B., Wu, M., Zhang, B.: Video question answering: a survey of models and datasets. Mob. Netw. Appl. pp. 1--34 (2021)

\bibitem{tang2023strucbench}
Tang, X., Zong, Y., Zhao, Y., Cohan, A., Gerstein, M.: Struc-bench: Are large language models really good at generating complex structured data? arXiv preprint arXiv:2309.08963  (2023)

\bibitem{touvron2023llama}
Touvron, H., Martin, L., Stone, K., Albert, P., Almahairi, A., Babaei, Y., Bashlykov, N., Batra, S., Bhargava, P., Bhosale, S., et~al.: Llama 2: Open foundation and fine-tuned chat models. arXiv preprint arXiv:2307.09288  (2023)

\bibitem{wang2023query2doc}
Wang, L., Yang, N., Wei, F.: Query2doc: Query expansion with large language models. In: EMNLP. pp. 9414--9423 (2023)

\bibitem{wang2020minilm}
Wang, W., Wei, F., Dong, L., Bao, H., Yang, N., Zhou, M.: Minilm: Deep self-attention distillation for task-agnostic compression of pre-trained transformers. In: NeurIPS. pp. 5776--5788 (2020)

\bibitem{wang2021tuta}
Wang, Z., Dong, H., Jia, R., Li, J., Fu, Z., Han, S., Zhang, D.: Tuta: Tree-based transformers for generally structured table pre-training. In: KDD. pp. 1780--1790 (2021)

\bibitem{wei2023chainofthought}
Wei, J., Wang, X., Schuurmans, D., Bosma, M., ichter, b., Xia, F., Chi, E., Le, Q.V., Zhou, D.: Chain-of-thought prompting elicits reasoning in large language models. In: NeurIPS. pp. 24824--24837 (2022)

\bibitem{whitehouse2023webie}
Whitehouse, C., Vania, C., Aji, A.F., Christodoulopoulos, C., Pierleoni, A.: {W}eb{IE}: Faithful and robust information extraction on the web. In: ACL. pp. 7734--7755 (2023)

\bibitem{wiseman-etal-2017-challenges}
Wiseman, S., Shieber, S., Rush, A.: Challenges in data-to-document generation. In: EMNLP. pp. 2253--2263 (2017)

\bibitem{wu2022texttotable}
Wu, X., Zhang, J., Li, H.: Text-to-table: A new way of information extraction. In: ACL. pp. 2518--2533 (2022)

\bibitem{yani2021challenges}
Yani, M., Krisnadhi, A.A.: Challenges, techniques, and trends of simple knowledge graph question answering: a survey. Information  \textbf{12}(7), ~271 (2021)

\bibitem{yu2023augmentation}
Yu, Z., Xiong, C., Yu, S., Liu, Z.: Augmentation-adapted retriever improves generalization of language models as generic plug-in. In: ACL. pp. 2421–--2436 (2023)

\bibitem{zhang2019bertscore}
Zhang, T., Kishore, V., Wu, F., Weinberger, K.Q., Artzi, Y.: Bertscore: Evaluating text generation with {BERT}. In: ICLR (2020)

\bibitem{zhong-etal-2022-video}
Zhong, Y., Ji, W., Xiao, J., Li, Y., Deng, W., Chua, T.S.: Video question answering: Datasets, algorithms and challenges. In: EMNLP. pp. 6439--6455 (2022)

\bibitem{zhong-chen-2021-frustratingly}
Zhong, Z., Chen, D.: A frustratingly easy approach for entity and relation extraction. In: NAACL. pp. 50--61 (2021)

\bibitem{zhou2022least}
Zhou, D., Sch{\"a}rli, N., Hou, L., Wei, J., Scales, N., Wang, X., Schuurmans, D., Cui, C., Bousquet, O., Le, Q., et~al.: Least-to-most prompting enables complex reasoning in large language models. arXiv preprint arXiv:2205.10625  (2022)

\bibitem{zhu-etal-2021-tat}
Zhu, F., Lei, W., Huang, Y., Wang, C., Zhang, S., Lv, J., Feng, F., Chua, T.S.: {TAT}-{QA}: A question answering benchmark on a hybrid of tabular and textual content in finance. In: ACL-IJCNLP. pp. 3277--3287 (2021)

\bibitem{zou2020survey}
Zou, Y., Xie, Q.: A survey on vqa: Datasets and approaches. In: CIT. pp. 289--297. IEEE (2020)

\end{thebibliography}

\end{document}